%% file: Goal_Oriented_chatbots_revised.tex
\documentclass[runningheads]{llncs}

\usepackage[T1]{fontenc}
\usepackage{graphicx}
\usepackage{amsmath}
\usepackage{amsfonts}
\usepackage{tikz}
\usepackage{wrapfig}
\usepackage{tablefootnote}
\usepackage{float}
\usepackage{multirow}
\usepackage{todonotes}
\usepackage{adjustbox}
\usepackage[hyperfootnotes=false]{hyperref}

\usetikzlibrary{positioning, arrows, calc}
\usepackage{color}

\newcommand{\tablefootnotemark}[1]{\textsuperscript{\getrefnumber{#1}}}
\newcommand\blfootnote[1]{%
  \begingroup
  \renewcommand\thefootnote{}\footnote{#1}%
  \addtocounter{footnote}{-1}%
  \endgroup
}
\begin{document}
\title{Towards Goal-Oriented Agents for Evolving Problems Observed via Conversation} 
\titlerunning{Agents for Problems Observed via Conversation}

\author{Michael Free\inst{1,3}\and
Andrew Langworthy\inst{1,2,3} \and
Mary Dimitropoulaki\inst{1,3}\and
Simon Thompson\inst{1}}
\authorrunning{M. Free et al.}
%
\institute{BT Applied Research, Ipswich, UK \and
Corresponding author: \email{andrew.langworthy@bt.com} \and
These authors contributed equally}

\maketitle              
\blfootnote{This preprint has not undergone peer review or any post-submission improvements or corrections. The Version of Record of this contribution is published in Artificial Intelligence XL. SGAI 2023. Lecture Notes in Computer Science, vol 14381, and is available online at \url{https://doi.org/10.1007/978-3-031-47994-6_11}.}
\begin{abstract}
The objective of this work is to train a chatbot capable of solving evolving problems through conversing with a user about a problem the chatbot cannot directly observe. The system consists of a virtual problem (in this case a simple game), a simulated user capable of answering natural language questions that can observe and perform actions on the problem, and a Deep Q-Network (DQN)-based chatbot architecture. The chatbot is trained with the goal of solving the problem through dialogue with the simulated user using reinforcement learning. The contributions of this paper are as follows: a proposed architecture to apply a conversational DQN-based agent to evolving problems, an exploration of training methods such as curriculum learning on model performance and the effect of modified reward functions in the case of increasing environment complexity.

\end{abstract}

\section{Introduction}
Task-oriented dialogue systems have found use in a wide variety of tasks. They have been employed successfully in several different applications where the task to be accomplished is well defined. Examples include restaurant booking \cite{lei-etal-2018-sequicity}, healthcare \cite{Athota20}, and entertainment \cite{li-etal-2017-end}. These tasks are often `data collection tasks' where a number of known information points must be collected by the agent to complete an end goal, with the information gathered rarely changing over the course of the conversation.

There are two main drawbacks with this approach, firstly that data-driven task-oriented dialogue systems require large amounts of heavily annotated training data, needing labels for intents, entities, and the flow from one dialogue state to another. This annotated data is often times consuming to produce, and in some scenarios does not exist. This can severely limit the ability to build and train such systems.

Secondly, in some tasks, particularly those where the agent is directing the user to perform some action, the changes applied by the user invalidate the information collected by the agent, and so require the beliefs of the agent to be updated. In contrast to many types of conversational problems, these evolving conversations require the agent to properly understand the problem that is being discussed and to be able to produce answers that cannot be the result of memorization. An example of such a task is an IT customer service agent attempting to diagnose and fix a fault (for example, with a router) for an end user. The actions the agent instructs the user to take (``Restart your machine'') can change the state of the problem, requiring the agent to update their information (``Is the light still blinking red?''). 

To study this problem, we introduce a `gridsworld' navigational game, where the aim is to move a square piece through a 2D maze to a goal. We introduce a simulated user, who can answer questions about the state of the game and, upon instruction, take actions on the gridsworld. Finally, we introduce an agent who can ask questions of the user and instruct the user to take actions on the gridsworld. Crucially, this agent cannot see or interact directly with the gridsworld environment; all interaction goes through the conversational intermediary.

The structure of this paper is as follows. In the next section we describe our goals, the datasets used, and the architectures of the models used during training. Next, we look at related work, in particular task-oriented chatbots, text-based games, and maze-solvers, in order to contextualise our contribution. Finally, we describe our experiments and discuss the results.

\section{Problem Formulation}
The goal of this work is to train a conversational agent to solve an evolving problem that it can only see and interact with via an intermediary.
\subsection{Overview of Approach}
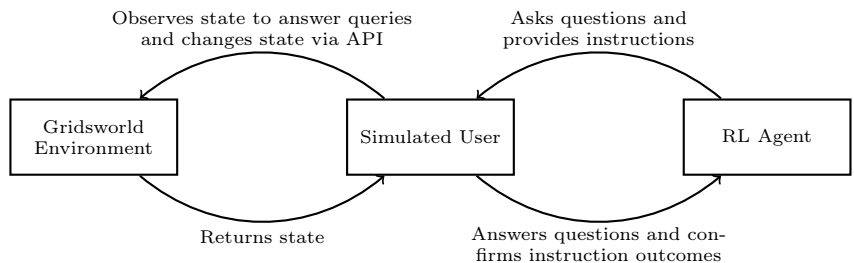
\begin{figure}
\centering
\begin{tikzpicture}[font=\scriptsize]
    \tikzset{block/.style={draw, thick, text width=2cm, minimum height=1cm, align=center}, 
             line/.style={-latex},
             arrowtext/.style={align=center, text width=4.5cm}}
    \node[block] (a) {Gridsworld Environment};
    \node[right=of a] (invis1) {};
    \node[block, right=of invis1] (b) {Simulated User};
    \node[right=of b] (invis2) {};
    \node[block, right=of invis2] (c) {RL Agent};

    \draw[->, thick] (b) edge [out=140, in=40] node[arrowtext, yshift=3mm] {Observes state to answer queries and changes state via API} (a);
    \draw[->, thick] (a) edge [out=320, in=220] node[arrowtext, yshift=-2mm] {Returns state} (b);
    \draw[->, thick] (b) edge [out=320, in=220] node[arrowtext, yshift=-3mm] {Answers questions and confirms instruction outcomes} (c);
    \draw[->, thick] (c) edge [out=140, in=40] node[arrowtext, yshift=3mm] {Asks questions and provides instructions} (b);  
\end{tikzpicture}
\caption{Problem architecture}\label{fig:problem_diag}
\end{figure}

The system we study consists of three components:
\begin{enumerate}
\item A gridsworld environment to be navigated by a reinforcement learning (RL) agent.
\item A simulated user, able to answer questions about and take actions in the gridsworld environment.
\item An RL agent to solve a problem in the gridsworld via the simulated user.
\end{enumerate}
These components interact with each other as shown in Figure \ref{fig:problem_diag}.

\subsection{Gridsworld Environment}
The gridsworld is a simple 2D environment that the simulated user can interact with at the request of the RL agent. It is a $6\times 6$ grid containing one circle and one square, between zero and ten impassable spaces (`obstacles'), and between zero and ten spaces that prevent the shapes from moving (`traps'). Appendix~\ref{app:1} shows an example gridsworld, with black spaces representing traps and light grey spaces representing obstacles. 


The goal of the agent is to move the square block through the grid, ignoring the obstacles and traps, to meet with the circle. At each step, the agent has a variety of actions to choose from. These are to move the square (e.g.\ ``Move the square to the right''), to ask a question about where the circle is in relation to the square (e.g.\ ``Is the square above the circle''), or to ask a question about where the nearest trap is (e.g.\ ``Where is the nearest trap?''). A full list of actions can be seen in Table~\ref{tab:utterances}.

If the agent attempts to move the square onto a space containing an obstacle or outside the grid boundaries, the move fails and the agent is informed. If the agent moves the square onto a space containing a trap, the next two attempts the agent makes to leave the trap fail, with the agent being informed that they are stuck in a trap.

We note that there are in the region of $10^{26}$ possible gridsworld states. While this is far fewer than some other possible problems classically solved using reinforcement learning techniques (the possible board states in a chess game, for example), there is still a considerable amount of complexity in the setup we introduce here. In addition, the problem is made even harder by the fact that the agent cannot observe the entire gridsworld environment; each question the agent asks only returns a small amount of information about the whole space. 

\subsubsection{Dataset Generation}
We generated $130{,}000$ gridsworld instances, with the placement of the blocks, traps, and obstacles chosen uniformly at random. For each instance, the number of obstacles and number of traps were also both chosen uniformly at random, with the number of each being between $0$ and $10$. Any scenes without a path between the circle and the square were discarded to make sure all the instances were solvable. Next, duplicates we removed (approximately $10\%$ of the instances) and the remainding instances were partitioned into an $80/10/10$ train/test/validation split. While this meant that the sets were disjoint, there may be some overlap in the conversational data that the agent trains and is tested on, where the solution is conversationally equivalent or a subset of a training conversation. For instance, the agent may be trained on a scene that involves moving the block 3 spaces to the left, but tested on the same scene with the target moved one space closer to the right. In this case, even though the scenes are different, the agent may be tested on a conversation which is a subset of a conversation seen in training.

\subsection{Simulated User}
The simulated user observes the gridsworld, can answer natural language questions about the environment, and can take actions upon request. This simulated user is based on the MAC network implementation from \cite{arad2018compositional}. The network is a combination of a convolutional neural network (CNN) acting as a feature extractor for the input image and a bidirectional long short-term memory (LSTM) to interpret the input questions. These two representations are then fed into a `MAC cell' to produce a natural language answer to the proposed questions.

Our MAC network is architecturally identical to that of Hudson and Manning, but we adapt the network to take in images of our gridsworld scenes rather than the CLEVR images of \cite{johnson2017clevr} the network was originally designed to interpret. 
 
Given a gridsworld scene we automatically generated questions, actions, and their associated answers in a rules-based fashion. The MAC network was then trained on these in an identical regime to that of \cite{arad2018compositional}.

\subsection{RL Agent}
The reinforcement learning agent converses with the simulated user, asking questions about and taking actions in the gridsworld. The simulated user can answer these questions and the RL agent then uses these answers to inform the next action. The RL agent cannot observe the gridsworld directly, it can gain information only through communication with the simulated user.

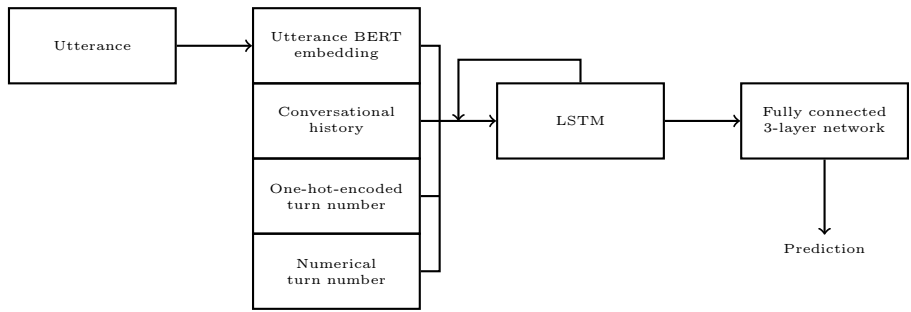
\begin{figure}
\centering
\begin{tikzpicture}[font=\tiny]
 \tikzset{block/.style={draw, thick, text width=2cm, minimum height=1cm, align=center}, 
             line/.style={-latex},
             arrowtext/.style={align=center, text width=4.5cm}}
    \node[block] (utt) {Utterance};
    \node[block, right=of utt] (bert) {Utterance BERT embedding};
    \node[block, node distance=-0.8pt, below=of bert] (conv) {Conversational history};
    \node[block, node distance=-0.8pt, below=of conv] (ohe) {One-hot-encoded turn number};
    \node[block, node distance=-0.8pt, below=of ohe] (numerical) {Numerical turn number};
    \node[block, right=of conv] (lstm) {LSTM};
    \node[block, right=of lstm] (dnn) {Fully connected 3-layer network};
    \node[below=of dnn] (preds) {Prediction};
    \draw[->, thick] (utt) edge (bert);
    \draw[->, thick] (conv) edge node[arrowtext, yshift=3mm] (input) {} (lstm);
    \draw[->, thick] let \p1=(input), \p2=(lstm.north) in (lstm.north) -- ++(0,0.3) -- ($(\x1,\y2)+(0,0.3)$)  -- ($(conv.east)!0.5!(lstm.west)$);
    \draw[->, thick] (lstm) edge (dnn);
    \draw[->, thick] (dnn) edge (preds);

    \draw[-, thick] let \p1=($(conv.east)!0.25!(lstm.west)$), \p2=(bert) in (bert) -- (\x1, \y2) -- (\p1);
    \draw[-, thick] let \p1=($(conv.east)!0.25!(lstm.west)$), \p2=(ohe) in (ohe) -- (\x1, \y2);
    \draw[-, thick] let \p1=($(conv.east)!0.25!(lstm.west)$), \p2=(numerical) in (numerical) -- (\x1, \y2) -- (\p1);

\end{tikzpicture}
\caption{Architecture of the RL agent}\label{fig:architecture}
\end{figure}

The architecture of the agent is shown in Figure~\ref{fig:architecture}. The agent takes as input the most recent utterance from the simulated user and concatenates it to the conversation history (all previous questions/actions/responses), as well as a numerical and one-hot-encoded counter, counting the size of the conversation histroy.

The conversation utterances are encoded as sentence-level embeddings using a pre-trained BERT model, introduced in \cite{reimers19}. In this work, all utterances from both the agent and user are created from templates and so they follow a fixed pattern, but further work could allow the simulated user to generate natural language without a template, see \cite{Li16}, and therefore an encoding method that could deal with this was chosen.

Once the input has been created, it is fed into an LSTM with $64$ hidden units. The output of that is then fed into a 3-layer network that returns a score over the $14$ next possible utterances for the agent. The first two layers have $32$ hidden units each.

The RL agent is trained using $Q$-learning \cite{watkins1989learning}. This is a reinforcement learning technique where a $Q$ function is learnt to assign a value to state-action pairs of a Markov decision process. The learning process is given by
\begin{equation}\label{eqn:Qlearning}
Q^\text{New}\left(s_t,a_t\right)\leftarrow r_t+\gamma\max_{a}Q\left(s_{t+1},a\right),
\end{equation}
where $r_t$ is the reward for taking the action $a_t$ at state $s_t$, and $\gamma$ is a discount constant. This function can be approximated by a neural network, as introduced in \cite{mnih13}, and that is the architecture our RL agent uses. The specific training regime is that of a double-DQN, where two $Q$-functions are learnt simultaneously, but in the update applied by Equation~\ref{eqn:Qlearning}, the functions are updated on each other. This solves some overestimation problems that Q-learning may have. For details, see \cite{hasselt16}. Specific details of the training methods are found in Section~\ref{sec:experiments}.

The code for our implementation of the RL agent is based on \cite{brenner18}, itself an instantiation of the code from \cite{li-etal-2017-end}.

\section{Related Work}
Combining a simulated user with a dialogue manager is a common technique in the literature to train task-oriented chatbots \cite{lei-etal-2018-sequicity,li-etal-2017-end}. These techniques often involve slot-filling for various tasks such as restaurant or movie theatre booking, and datasets containing these sorts of problems are well known, for example various iterations of the MultiWoZ dataset \cite{budzianowski18,zang20}. Our problem space differs from these in that the information gained from asking a question about the gridsworld changes over time as the square moves around the environment and it's relationship to other objects changes. More recent work in chatbots has experimented with a variety of more complex scenarios, for example negotiation \cite{verma22}, but the same limitations still remain, with information being largely static. 
 
Text-based games and the models built to play them, such as those studied in \cite{textgames1,textgames2,textgames3}, offer a richer problem space. In particular, in \cite{Yuan18} a DQN-based agent is used to solve a text-based game to navigate a series of rooms. Here the problem evolves over the course of the agent instructions and the agent must remember previous navigation instructions and the results to obtain high performance. However, the agent does not have the ability to query the game or otherwise gain information without taking an action, which renders it different to our approach.

Solving navigational problems similar to the gridsworld we introduce is a natural problem upon which to apply reinforcement learning techniques. Examples include an RL agent designed to navigate a maze built in Minecraft \cite{frazier19}. The RL agent is augmented by a synthetic oracle, that gives one-hot-encoded directional advice to the agent. In related work, Zambaldi et al.\ \cite{zambaldi18} require the agent to travel to various locations within a 2D box-world before completing the task. They achieve this by using relational reinforcement learning, where the relations are learned using a self-attention mechanism. This allows the agent to make long term plans about its route. Building from this prior work, the relational grid world is introduced in \cite{kucuk21}, where an agent should navigate the 2D world to get to an exit point. This world contains various objects including walls and mountains, that affect the movements that the agent can make, and mean that the optimal path may not be the shortest. These all differ from our approach in that the agent trained can all directly act upon and (at least partially) observe the world, whereas our problem formulation requires information to be gained via an intermediary.

\section{Experiments}\label{sec:experiments}
The RL agent was trained on the navigation task in a number of scenarios. The neural network was updated using the Bellman equation, Equation~\ref{eqn:Qlearning}, where the states consist of the conversational histories and the actions are the possible actions the agent can take (either ask a question or request the square moved). We set $\gamma=0.9$. The reward function for the agent was deliberately kept simple, since a future goal of this research is to investigate whether the learning and approaches made in this fairly easy game can be transferred to a more complex domain. The agent was given a reward of $-1$ for each action it took, unless the action completed the gridsworld scenario, i.e.\ moved the square into the same place as the circle, in which case the reward given was $60$. The small negative rewards encourage efficient completion of the scenario. A limit of $30$ turns was given to each scenario, and if the agent failed to complete the task within that time, an extra penalty of $-30$ was added.

The agent used an $\varepsilon$-greedy policy with slow annealing. We initialised $\varepsilon=0.2$, and annealled it so that $\varepsilon=0.01$ (i.e.\ each action taken by the agent has a $1\%$ chance of being chosen at random) after $1.15\text{m}$ training episodes. The replay memory size was set to $51200$, and the model was updated using a mini-batch size of $512$. The optimiser used was Adam with a learning rate of 0.0001.

\subsection{Curriculum Learning}
To speed up training times, we applied a curriculum learning regime \cite{bengio09} to the model. Before training, we partitioned the training data into three sets based on the minimum number of turns required to complete the scenario using an implementation of Dijkstra's algorithm. The three partitions of the training data were given by path lengths of less than $4$, path lengths of between $4$ and $5$ inclusive, and paths of length greater than $5$. These numbers include any potential time where the agent is stuck in a trap. The short scenarios made up approximately $40\%$ of the training data, and the medium and long scenarios made up $30\%$ each of the data.

During training, the model learned in a guided way, where at first only the problems with short path lengths are seen. Once the model reached an average reward greater than $10$ over $500$ scenarios, problems with medium length paths were added in. This is repeated once more, where all scenes were added to the training data. Since the model only got a positive reward upon completion of the scenario, this curriculum learning structure dramatically decreased learning time since it was far easier to get rewards early on in training.

\begin{figure}
    \begin{minipage}[c]{0.5\linewidth}
        \centering
        \includegraphics[scale=0.4]{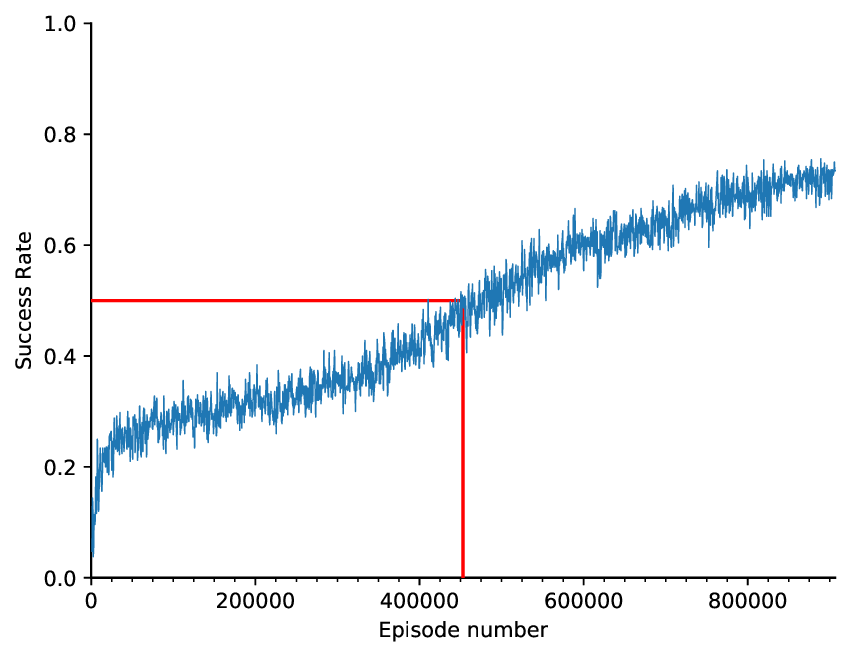}
        \caption{No curriculum learning} \label{fig:no_curr}
    \end{minipage}\hfill
    \begin{minipage}[c]{0.5\linewidth}
        \centering
        \includegraphics[scale=0.4]{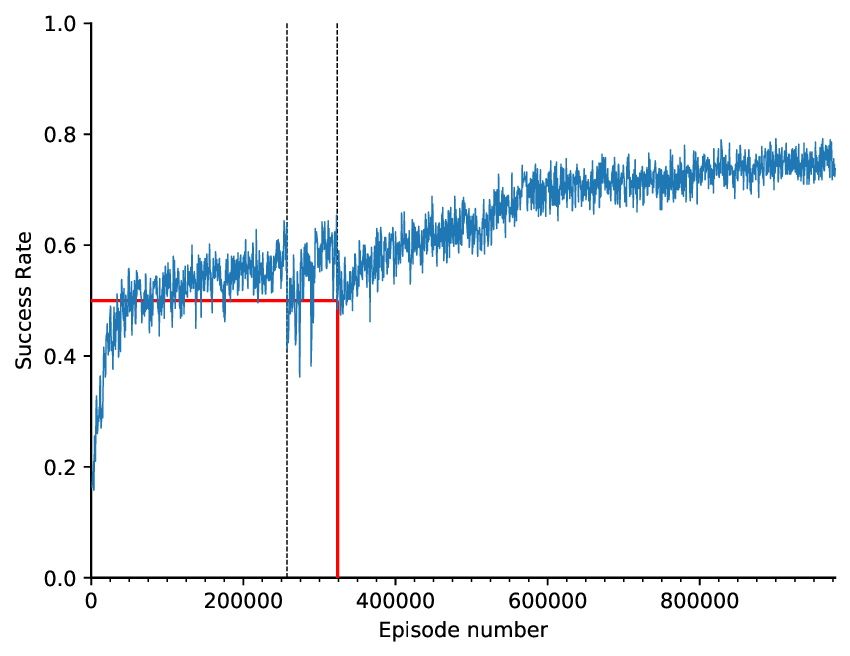}
        \caption{Curriculum learning} \label{fig:curr}
    \end{minipage}
\end{figure}

In Figures~\ref{fig:no_curr} and \ref{fig:curr} we see two training runs that highlight the difference between curriculum learning versus no curriculum learning respectively. We can see the hard drops in success rate denoted by the vertical black dashed lines in Figure~\ref{fig:curr}, which correspond to the addition of medium and hard scenes into the training data. The performance graph of Figure~\ref{fig:no_curr} by contrast is comparatively smooth. The agent trained using a curriculum learning regime reached a success rate of $0.5$ on the full dataset by $324{,}000$ runs, whereas it took the agent trained on the whole dataset from the beginning $453{,}000$ runs, a $39.8\%$ increase. These spots are marked on the graphs with red lines. 

\subsection{Alternative Agent Architectures}\label{sec:architectures}
During testing, we experimented with several architectures for the RL agent. These followed the same structure as in Figure~\ref{fig:architecture} with the difference that we replaced the LSTM unit a different one. For the first variant, the DNN-based architecture, we replaced the LSTM module by a fully connected layer with $64$ hidden units. For the second variant, we called the CNN-based architecture, we replaced the LSTM module by two one-dimensional convolutional layers with a pooling layer of pool size $2$ after each one. The first CNN layer had $64$ filters and the second had $32$. Both had a kernel size of $3$. They were trained in the same way as the LSTM-based architecture, with identical hyperparameter choices.

While an LSTM-based architecture is a classic way to deal with problems such as our conversational traversal of the gridworld due to the nature of the information gathered at earlier timesteps becoming less useful as more utterances are made, we tested the other architectures for comparison's sake. 

\subsection{Modified Rewards}
During experiments, we modified the reward function to improve agent performance. We did this by giving the agent a small reward for asking questions to encourage question asking early in training, and for taking a move that moved the square closer to the circle. The introduction of these rewards marginally sped up the training time but did not result in a higher average success rate, implying that even without these extra rewards, the agent was settling on a policy that valued asking questions and moving closer to the target.

\section{Results}
\begin{figure}
    \begin{minipage}[c]{0.5\linewidth}
        \centering
        \includegraphics[scale=0.4]{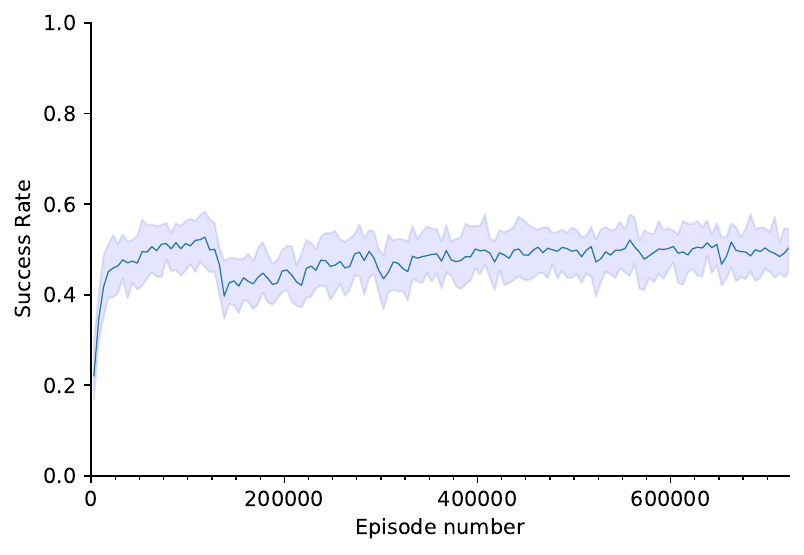}
        \caption{DNN-based architecture} \label{fig:dnn}
    \end{minipage}\hfill
    \begin{minipage}[c]{0.5\linewidth}
        \centering
        \includegraphics[scale=0.4]{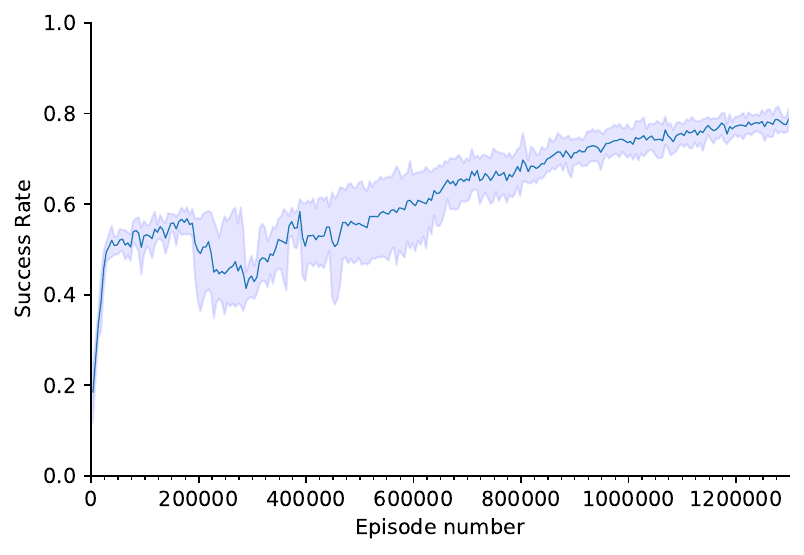}
        \caption{LSTM-based architecture} \label{fig:lstm_eps}
    \end{minipage}
\end{figure}

Table~\ref{tab:results} shows the success rate of various agent architectures when trained to solve the gridsworld problems. The difference in architecture is only in the LSTM layer taking the encoded conversational context and turn numbers as described in Section~\ref{sec:architectures}. All results shown below apart from the human baseline are the average of three different tests on a hold-out dataset. Figures~\ref{fig:dnn} and \ref{fig:lstm_eps} show the training performance of the DNN and LSTM-based architectures over the three runs, with the solid blue lines denoting the average of the three.

\begin{table}
\caption{Success rates of different RL agent architecture}\label{tab:results}
\centering
\begin{tabular}{ll}
\hline
\textbf{RL agent architecture} &  \textbf{Success rate}\\
\hline\hline
Human Baseline &  \textbf{0.95}\tablefootnote{Human baseline obtained a lower average reward than the LSTM-based architectures}\\
Fully connected layers &  0.49\tablefootnote{\label{note:thresh}Agents did not reach a high enough average reward for the threshold to add in the hardest $30\%$ of scenes}\\
CNN & 0.58\tablefootnotemark{note:thresh}\\
LSTM & \textbf{0.79}\\
LSTM - modified reward & \textbf{0.79}\\\hline
\end{tabular}
\end{table}

\subsection{Human Baseline}
As a baseline, we ran human tests on $50$ episodes. The success rate was $95\%$, with an average reward of $43.4$. Although this success rate is considerably higher than that of the highest performing agent, we note that the average reward is actually lower. A partial explanation for this result is probably non-optimal questions and movement from the human operator; factors which we would expect the agent to perform better at than a human. However, we conjecture that the main factor is that the agent is optimising based on the reward it is getting, not the successful outcome. This might mean that it makes `riskier' moves, since this gives a better average reward, rather than playing safe to ensure a success.

\subsection{DNN Network}
Figure~\ref{fig:dnn} shows the results of the fully connected network without an LSTM layer. This model stopped learning quickly. In testing, it achieved a success rate of $49.2\%$, but it did not achieve a high enough reward to reach the threshold needed to add in the hardest $30\%$ of scenes. This shows the difficulty that the fully connected network has in effectively encoding the long-term history of the conversation. It is possible that better results could have been achieved with a deeper network.

\subsection{LSTM Network}
The trained agent, whose results are in Figure~\ref{fig:lstm_eps}, achieved an episode completion rate of $84\%$, with an average reward of $44.6$ out of a maximum $60$. The average reward for successful episodes (considering uncompleted episodes got a total reward of $-60$) was $54$. Of the conversational turns taken, $61.6\%$ of them were actions and $39.4\%$ were questions, with $5\%$ of those relating to the whereabouts of the nearest trap.

An interesting point to note about these graphs is the period of slow learning after the initial rapid improvement of the system (see Figures~\ref{fig:no_curr}, \ref{fig:curr}, \ref{fig:lstm_eps}). Since any action by the agent carries the same negative reward but only the movement actions have a chance of getting the large reward for completing the task, asking questions is discouraged for a naïve policy. Also, the answer to a question must then be understood in terms of how it relates to the environment. The combination of these factors means that the agent initially learns a suboptimal strategy of exploring without asking questions. Only once the state space is sufficiently explored does the agent's strategy shift to using questions.

This may also explain the behaviour of the agent shown in Figure~\ref{fig:no_curr}. After a period of slow improvement success rate the gradient of the performance curve changes and the agents start to improve faster. We hypothesise that this is point where the agent learns how to parse the information given by asking questions, and so the optimal policy rapidly shifts.

\section{Conclusion}
We have shown that the proposed system architecture, a double DQN-based dialogue agent, can be trained to solve evolving problems that it can only observe via conversation with a simulated user. Through this conversation, it can achieve a high degree of success. The simulated user is also demonstrated to be suitable for incorporation into the described framework.

We have demonstrated that by utilising curriculum learning -- ordering the training epochs in increasing scene complexity -- we need $40\%$ fewer training scenes to reach the same performance – a large gain when considering the resource intensive nature of training RL models.

Exploration of a modified reward function to account for environmental complexity, such as explicit negative reward for traps, showed little performance increase over a reward based solely on task completion metrics such as success and turns taken.

\subsection{Transferability to Real Problems and Future Work}
Whilst we have shown the agent is able to solve problems conversationally in the framework described, there are still shortcomings when it comes to transferring to a real problem. The simulated user is limited to a small number of conversational options, which is not a limitation placed on real users. This in turn means our trained reinforcement learning agent is only trained to deal with those specific utterances. In future work a generative model could be utilised to generate a wider range of conversational responses. 

A positive aspect of the system is the non-perfection of the simulated user. As the trained model occasionally makes mistakes, the RL agent learns robustness to individual incorrect answers.

The model also exhibits behaviour that is optimal in the context of the reward function given, but may not be suited to real interaction, such as abandoning questions and utilising only movement in a searching pattern as the conversational turn limit is approached. Further exploration of the reward function optimal for customer experience will need to be explored.

To transfer to a real-world problem, a simulation of that problem, along with a capable simulated user would also need to be constructed. We can think of these in terms of numbers of degrees of freedom that we can add to the system. We have multiple options here, the grid size or shape could be upgraded or changed; the number of question iterations can be changed; the reward function could be tweaked to better represent the problem; we could even introduce new elements to the world, taking inspiration from constructions like \cite{kucuk21}. These increases in complexity of the problem could mirror the increase in complexity inherent in real-world problems. Very complex scenarios such as those involving out-of-domain elements (such as other items in the customer's home interfering with the operation of equipment) would be out of scope for such a simulation (and therefore the agent) due to the complexity of simulation required.

\subsubsection{Acknowledgements}
The authors would like to thank Rob Claxton for his help and guidance.



%
%
%

\bibliographystyle{splncs04}
\bibliography{paper_refs}

\include{appendices}
\end{document}

%% file: appendices.tex
\appendix
\section{Example Conversation}\label{app:1}
The following is an example conversation recorded between the RL agent and the simulated user.
    \begin{enumerate}
    \item Is the square above the circle?\\
    \emph{Yes}
    \item\label{conv:learn} Is the square left of the circle?\\
    \emph{No}
    \item Is the square above the circle?\\
    \emph{Yes}
    \item Move the square down.\\
    \emph{Yes}
    \item Is the square above the circle?\\
    \emph{Yes}
    \item Move the square down.\\
    \emph{Yes}
    \item Is the square above the circle?\\
    \emph{No}
    \item\label{conv:apply} Move the square left.\\
    \emph{Yes}
    \item Move the square left.\\
    \emph{Yes}
    \item Move the square left.\\
    \emph{Yes}
    \item Move the square left.\\
    \emph{Complete}
    \end{enumerate}
    \begin{figure}[H]
    \centering 
    \scalebox{0.6}{
    \begin{tikzpicture}
        \foreach \trap in {(1,0), (5,1), (3,2), (2,4), (1,5), (3,5)}
            {\fill[black] \trap rectangle +(1,1);}
        \foreach \wall in {(5,4), (3,4), (2,5), (1,3), (0,0), (0,3)}
            {\fill[gray] \wall rectangle +(1,1);}
        \draw[fill=red] (2.5, 3.5) circle (0.4);
        \draw[fill=blue] (4.15, 5.15) rectangle +(0.7, 0.7);
        \draw[black, thick] (0,0) grid (6,6);
    \end{tikzpicture}}
    \caption{Initial gridsworld for the example conversation} \label{fig:gridsworld_ex1}
    \end{figure}
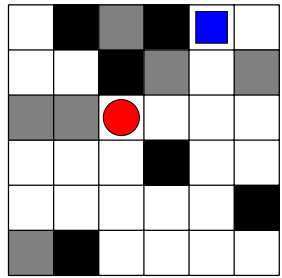

Note that in \ref{conv:learn}) the agent learns that the square is not to the left of the circle, and this knowledge is retained over the next few call and response pairs, since in \ref{conv:apply}) the agent immediately knows to start moving left rather than confirm the relative position of the square.
\newpage
\section{Table of Utterances}
Table~\ref{tab:utterances} contains all the possible utterances of both the RL agent and the simulated user.

\begin{table}
\addtolength{\leftskip}{-30pt}
\addtolength{\rightskip}{-30pt}
\newcommand\firstcol{0.1\textwidth}
\newcommand\secondcol{0.2\textwidth}
\newcommand\thirdcol{0.3\textwidth}
\newcommand\fourthcol{0.4\textwidth}
\caption{User/RL agent utterances}\label{tab:utterances}
\begin{tabular}{p{\firstcol}p{\secondcol}p{\thirdcol}p{\fourthcol}}
\textbf{Speaker} & \textbf{Utterance Type} & \textbf{Text} & \textbf{Notes}\\
\hline\hline
\multirow[t]{9}{\firstcol}{RL Agent} & \multirow[t]{4}{\secondcol}{Relational Question} & Is the square above the circle? & \multirow[t]{9}{*}{}\\\cline{3-3}
                          &                                     & Is the square below the circle? &                 \\\cline{3-3}
                          &                                     & Is the square to the right of the circle? &       \\\cline{3-3}
                          &                                     & Is the square to the left of the circle? &        \\\cline{2-3}
                          & Trap Question                       & Where is the nearest trap? &                      \\\cline{2-3}
                          & \multirow[t]{4}{\secondcol}{Movement Command}    & Move the square up                       &        \\\cline{3-3}
                          &                                     & Move the square down                       &      \\\cline{3-3}
                          &                                     & Move the square to the left              &        \\\cline{3-3}
                          &                                     & Move the square to the right             &        \\\hline

\multirow[t]{8}{\firstcol}{Simulated User} & \multirow[t]{2}{\secondcol}{Relational Question Response} & Yes & \multirow[t]{2}{*}{}                 \\[5pt]\cline{3-3}
                                &                                               & No  &                                   \\[5pt]\cline{2-4}
                                & \multirow[t]{2}{\secondcol}{Movement Command Response}    & Yes & \multirow[t]{2}{\fourthcol}{The answer tells the agent
                                                                                                        whether the movement was 
                                                                                                        successful or not}\\[5pt]\cline{3-3}
                                &                                               & No  &                                   \\[5pt]\cline{2-4}
                                & \multirow[t]{4}{\secondcol}{Trap Question Response}       & There are no traps in the scene            &                  \\\cline{3-4}
                                &                                               & It is $X$ moves $D_1$ and $Y$ moves $D_2$  &  $X$ and $Y$ are integers between $1$ and $5$,
                                                                                                                                $D_1$ and $D_2$ are directions: up, down,
                                                                                                                                left, or right  \\\cline{3-4}
                                &                                               & It is $X$ moves $D_1$                      &  $X$ is an integer between $1$ and $5$,
                                                                                                                                $D_1$ is a direction: up, down,
                                                                                                                                left, or right  \\\cline{3-4}
                                &                                               & You're in one!  &  \\

\end{tabular}
\end{table}